\documentclass[journal]{IEEEtran}
\usepackage{amsmath,amsfonts}
\usepackage{algorithmic}
\usepackage{algorithm}
\usepackage{array}
\usepackage[caption=false,font=normalsize,labelfont=sf,textfont=sf]{subfig}
\usepackage{textcomp}
\usepackage{stfloats}
\usepackage{url}
\usepackage{verbatim}
\usepackage{graphicx}
\usepackage{cite}
\usepackage{booktabs}
\usepackage{float}
\usepackage{placeins}
\usepackage{multirow}
\usepackage{xcolor}
\usepackage{colortbl}
\usepackage{pifont}
\newcommand{\cmark}{\ding{51}}
\newcommand{\xmark}{\ding{55}}
\begin{document}


\title{Agentic AI-based Framework for Mitigating Premature Diagnostic Handoff and Silent Hallucination in Healthcare Applications}

\author{
\IEEEauthorblockN{Divyansh Srivastava\textsuperscript{1,2},
Shreya Ghosh\textsuperscript{2}, Anshul Verma\textsuperscript{1, 3}
Rajkumar Buyya\textsuperscript{1}}

\IEEEauthorblockA{\textsuperscript{1}Quantum Cloud Computing and Distributed Systems (qCLOUDS) Lab,\\
School of Computing and Information Systems,\\
The University of Melbourne, Australia}

\IEEEauthorblockA{\textsuperscript{2}Department of Computer Science and Engineering,\\
School of Electrical and Computer Sciences (SECS),\\
Indian Institute of Technology Bhubaneswar, India}

\IEEEauthorblockA{\textsuperscript{3}Department of Computer Science\\  
Banaras Hindu University, Varanasi, India}
}

\maketitle

\begin{abstract}
Recent advances in Large Language Models (LLMs) and multi-agent systems have driven the rise of Agentic AI, showing promise for medical reasoning. However, open-ended conversational agents remain prone to two critical failure modes: premature diagnostic handoff and silent clinical hallucinations that may go undetected before reaching the patient. In this work, we propose a multi-agent framework that addresses both issues by replacing ``LLM-as-a-judge'' routing with deterministic orchestration constraints. The framework incorporates two safety mechanisms. First, a neuro-symbolic state-tracking gate enforces completeness of the OLDCARTS clinical protocol (Onset, Location, Duration, Character, Aggravating/Alleviating factors, Radiation, Timing, and Severity) by blocking diagnostic transitions until all required dimensions are collected. Second, an epistemic uncertainty quantification (UQ) gate computes semantic entropy ($H$) across $K=5$ independent diagnostic samples to identify and intercept divergent outputs before delivery.

We evaluate the system using simulated patient agents powered by the llama-3.1-70b-instruct model on 150 test cases. The full architecture achieves $49.3\%$ diagnostic precision, representing an absolute improvement of $11.3$ percentage points over an unconstrained baseline. Additionally, we observe a statistically significant negative correlation ($r = -0.181,\ p < 0.05$) between OLDCARTS completeness ($\sigma$) and semantic entropy ($H$), suggesting that structured information gathering is associated with reduced diagnostic uncertainty.
\end{abstract}



\section{Introduction}
\IEEEPARstart{C}{linical} triage and diagnosis are high-stakes processes where errors in history taking or diagnostic reasoning can lead to delayed treatment, patient harm, and loss of trust in AI-assisted systems~\cite{lu2024triageagent}. With increasing demand on healthcare systems and persistent workforce shortages, there is growing interest in deploying AI-based clinical decision support tools~\cite{srinivasu2026agentic}. Large Language Models (LLMs) have emerged as promising candidates due to their ability to encode medical knowledge, perform reasoning, and interact with patients through natural language.

Recent work demonstrates the potential of LLMs in clinical applications. Generalist models such as MedFound support diagnosis across multiple specialties~\cite{liu2025generalist}, while adapted LLMs have shown strong performance in structured clinical summarization tasks~\cite{vanveen2024summarization}. Specialized systems such as RadGPT further highlight the capability of LLMs to generate patient-specific explanations~\cite{herwald2025radgpt}. These advances suggest a pathway toward conversational agents capable of assisting in end-to-end clinical workflows.

However, reliability remains a critical challenge. Medical LLMs are prone to hallucination, producing incorrect diagnoses, fabricated medications, or unsafe recommendations~\cite{omar2025misinformation}. Such failures are not rare and have been observed across both general-purpose and domain-specific models. A key limitation is that existing approaches optimize for fluent generation rather than epistemic correctness.

\begin{figure}
    \centering
    \includegraphics[width=0.8\linewidth]{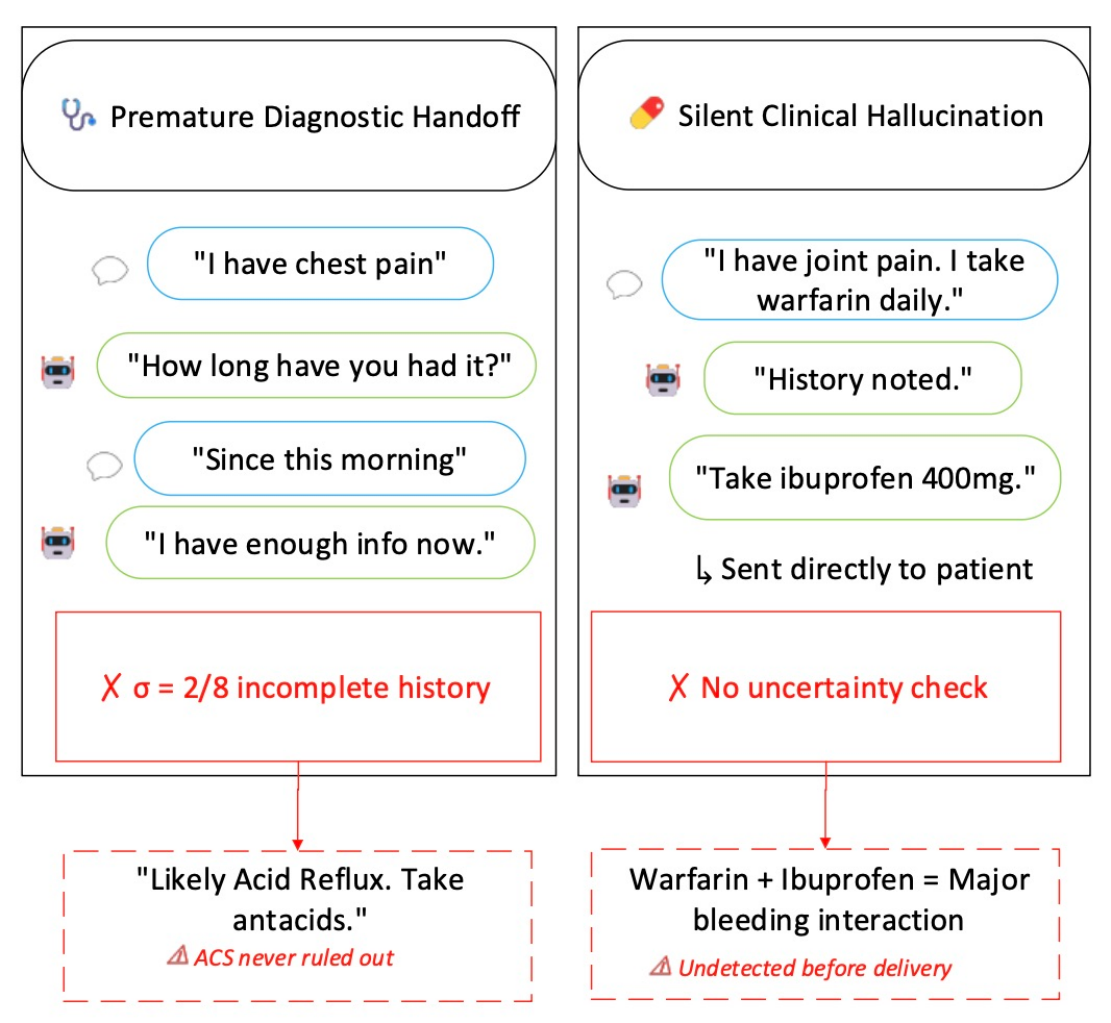}
    \caption{ Illustration of two critical failure modes 
of unconstrained LLM clinical agents: premature 
diagnostic handoff and silent hallucination.}
    \label{fig:motivation}
\end{figure}

Current mitigation strategies, including retrieval-augmented generation, chain-of-thought prompting, and LLM-as-a-judge frameworks, operate in a probabilistic manner. While they can reduce error rates, they do not provide guarantees of correctness. More importantly, they do not ensure \textit{intake completeness}, i.e., whether all clinically required symptom dimensions are collected before diagnosis. In clinical practice, structured history taking using the OLDCARTS protocol is essential, as incomplete information is a major source of diagnostic error~\cite{bickley2012bates}. Existing multi-agent systems do not enforce this requirement through a deterministic mechanism. Figure~\ref{fig:motivation} provides a concrete illustration of the two failure modes that motivate this work. The left panel depicts \emph{premature diagnostic handoff}, where the agent proceeds to diagnosis after collecting only a partial symptom history, with just $2$ of the $8$ OLDCARTS dimensions observed. The resulting diagnosis is therefore based on incomplete clinical context. The right panel depicts \emph{silent clinical hallucination}, where the agent produces a medication recommendation without any explicit uncertainty or safety screening, despite a potentially serious drug interaction. Taken together, these scenarios show that reliability in clinical conversational systems depends not only on the quality of the generated diagnosis, but also on whether the system enforces structured intake and flags uncertain or potentially unsafe outputs before delivery.

In this work, we address these limitations through a neuro-symbolic multi-agent framework 
where we replace purely prompt-based control with deterministic orchestration at the system level. Specifically, we introduce two complementary mechanisms. First, a Neuro-Symbolic OLDCARTS State Tracker (M1), which enforces completeness of symptom collection by blocking transition to diagnosis until all required fields are observed. Second, a Semantic Entropy-based Uncertainty Quantification (UQ) gate (M2), that computes normalized entropy across different independent diagnostic samples to identify divergent outputs before they are presented to the user.

We evaluate the proposed framework on simulated clinical cases derived from MedQA~\cite{jin2021medqa}, using LLM-based patient agents realised using Agentic-AI software system. Results show that the proposed system improves diagnostic accuracy by $11.3$ percentage points over an unconstrained baseline. Additionally, we observe a statistically significant negative correlation between symptom completeness and diagnostic uncertainty, indicating that structured information collection contributes to more consistent model outputs.

The main contributions of this paper are summarized as follows:
\begin{itemize}
    \item \textbf{Neuro-symbolic multi-agent framework:}  
    We proposed a structured multi-agent architecture, based on Agentic AI paradigm, for clinical triage and diagnosis that separates \textit{history taking}, \textit{diagnostic reasoning}, and \textit{safety supervision} into distinct role-specialized components. By organizing the workflow in this manner, the framework supports explicit control over the transition from symptom intake to diagnosis, rather than relying solely on unconstrained conversational generation.

    \item \textbf{Deterministic OLDCARTS-based intake verification mechanism:}  
    We introduce a neuro-symbolic state-tracking gate that explicitly monitors the completeness of symptom collection under the OLDCARTS clinical protocol. This mechanism enforces a deterministic verification step before diagnosis is initiated, thereby reducing the risk of premature diagnostic handoff caused by incomplete history taking.

    \item \textbf{Uncertainty-aware diagnostic screening mechanism:}  
    We incorporate a semantic entropy-based uncertainty quantification module that evaluates disagreement across multiple independently generated diagnostic samples. This mechanism provides an additional screening layer for identifying potentially unreliable or divergent diagnostic outputs before they are presented to the user.

    \item We evaluate the proposed framework through an ablation study across multiple test settings and show that the our proposed architecture improves diagnostic accuracy ($+11.3$ percentage-point 
    accuracy gain) over an unconstrained baseline. We further analyze the relationship between structured symptom completeness and diagnostic uncertainty, showing that more complete intake is associated with lower entropy in downstream diagnosis generation.
\end{itemize}

\begin{figure*}[t]
    \centering
    \includegraphics[width=\textwidth]{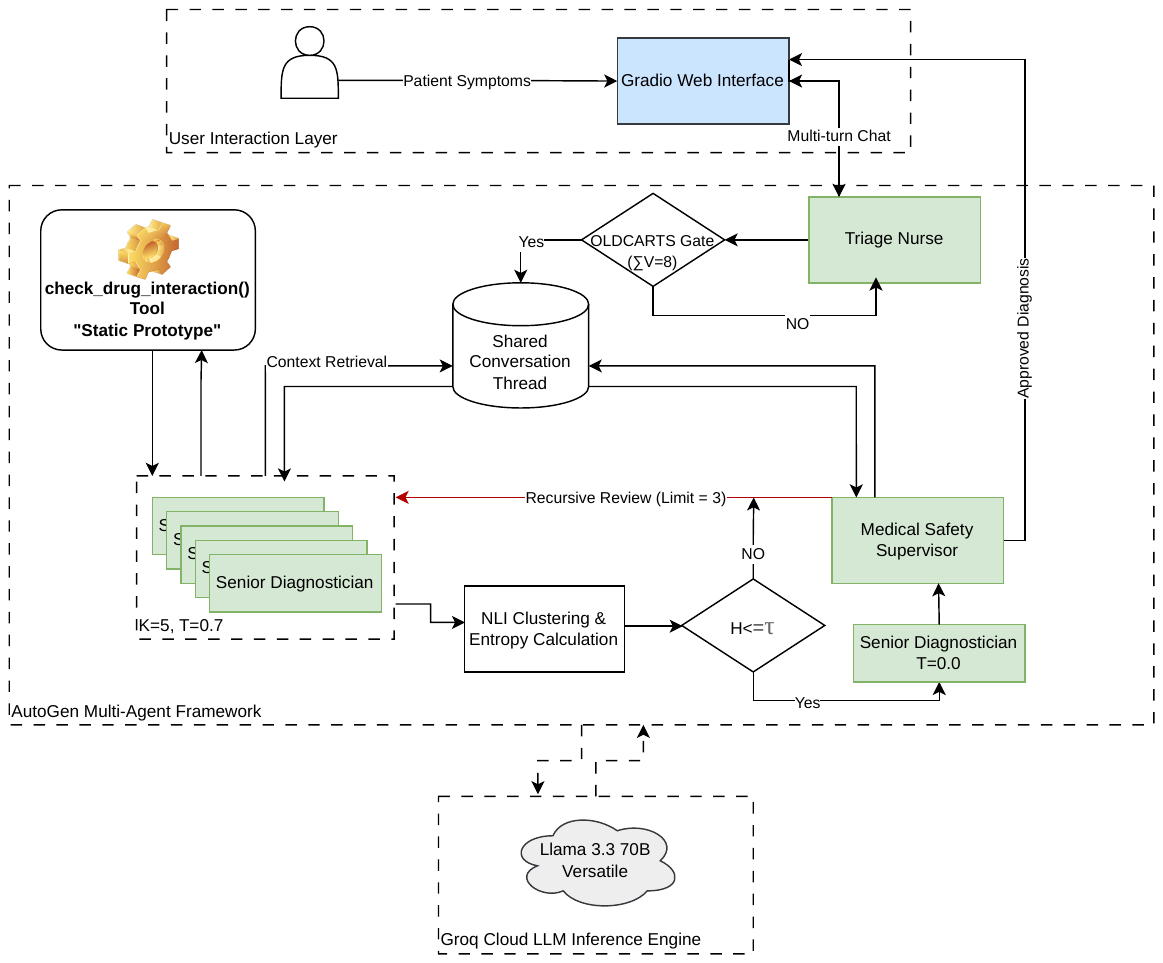}
    \caption{System architecture of the proposed Neuro-Symbolic Multi-Agent 
    Triage framework. The pipeline operates across three phases: (1) structured 
    history taking enforced by the OLDCARTS state-tracking gate [M1], (2) 
    parallel stochastic diagnostic sampling with semantic entropy uncertainty 
    quantification [M2], and (3) a recursive safety supervision loop with up to 
    three revision attempts.}
    \label{fig:architecture}
\end{figure*}


\begin{table*}[t]
\centering
\caption{Comparison of Related Multi-Agent and LLM-Based Clinical Frameworks}
\label{tab:related}
\renewcommand{\arraystretch}{1.3}
\begin{tabular}{lcccccc}
\toprule
\textbf{System} & \textbf{Multi-Agent} & \textbf{Structured Intake} & \textbf{Formal Gate} & \textbf{UQ / Hallucination} & \textbf{Safety Supervision} & \textbf{Open-Source LLM} \\
\midrule
Med-PaLM 2 \cite{singhal2023medpalm2}       & \xmark & \xmark & \xmark & \xmark & \xmark & \xmark \\
GPT-4 (USMLE) \cite{nori2023capabilities}         & \xmark & \xmark & \xmark & \xmark & \xmark & \xmark \\
Conv. Health Agents \cite{abbasian2023conversational} & \cmark & Partial & \xmark & \xmark & \xmark & \cmark \\
LLM-as-a-Judge \cite{zheng2023judging}       & \cmark & \xmark & \xmark & \xmark & Probabilistic & \xmark \\
\textbf{Proposed (Ours)}                      & \cmark & OLDCARTS & \textbf{Symbolic} & \textbf{SE Gate [M2]} & \textbf{Recursive} & \cmark \\
\bottomrule
\end{tabular}
\end{table*}

\section{Related Work}
\label{sec:related}

\textbf{LLMs in Clinical Decision Support.} The use of LLMs in clinical decision support has expanded rapidly with the emergence of instruction-tuned foundation models. Early evidence of this potential came from studies showing that GPT-4 could perform strongly on the United States Medical Licensing Examination (USMLE) without task-specific fine-tuning, highlighting its broad medical knowledge and reasoning capability~\cite{nori2023capabilities}. Subsequent domain-adapted models, such as Med-PaLM 2~\cite{singhal2023medpalm2}, further narrowed the gap with specialist-level performance on medical question answering benchmarks. More recently, generalist medical models including MedFound~\cite{liu2025generalist} have demonstrated the feasibility of supporting diagnosis across multiple clinical specialties within a unified model family. In parallel, specialized applications such as RadGPT for radiology report explanation~\cite{herwald2025radgpt} and adapted LLMs for clinical dialogue summarization~\cite{vanveen2024summarization} have shown that these models can achieve strong performance on targeted clinical language tasks, including settings where they match or surpass human experts in structured information extraction.

\textbf{Hallucination and Safety in Medical AI.}  
Despite the growing capabilities of LLMs in healthcare, hallucination remains a major unresolved challenge. Omar et al.~\cite{omar2025misinformation} show that LLMs can absorb and reproduce clinical inaccuracies embedded in realistic patient narratives, often treating false medical context as valid evidence. To mitigate such failures, prior work has explored strategies such as retrieval-augmented generation (RAG), chain-of-thought prompting, and self-consistency decoding. While these methods can reduce hallucination rates, they remain inherently probabilistic and do not provide deterministic safeguards. Similarly, the ``LLM-as-a-judge'' paradigm, in which a secondary LLM evaluates the output of a primary model~\cite{zheng2023judging}, can improve output quality on average, but it does not eliminate failure modes in a guaranteed manner, since the judging model is itself susceptible to the same underlying errors.

\textbf{Multi-Agent Systems for Healthcare.} Multi-agent frameworks offer a structural mechanism for decomposing complex clinical workflows into role-specialized components. Conversational Health Agents \cite{abbasian2023conversational} propose a personalized LLM-powered agent framework that separates symptom elicitation from downstream planning. The Microsoft's AutoGen agentic AI framework \cite{autogen2024} provides a general-purpose multi-agent conversation infrastructure that has been applied to medical question answering and summarization tasks. However, existing multi-agent healthcare systems typically rely on prompt-based constraints for inter-agent routing, leaving the completeness of information gathering to the model's instruction-following probability rather than verified by a formal symbolic gate.

\textbf{Uncertainty Quantification in LLM Outputs.}
Uncertainty quantification (UQ) for LLMs has emerged as a distinct research area. Token-probability-based measures suffer from the ``linguistic paraphrase problem'': semantically equivalent outputs (e.g., ``STEMI'' vs. ``ST-elevation myocardial infarction'') produce artificially high entropy. Semantic Entropy (SE) \cite{kuhn2023semantic} addresses this by clustering semantically equivalent outputs via bidirectional NLI entailment before computing normalized Shannon entropy. This estimator has been validated on open-domain QA tasks but has not previously been applied to the multi-label, multi-diagnosis setting of clinical differential diagnosis or integrated into a deterministic multi-agent orchestration gate.


Table~\ref{tab:related} compares the proposed framework with representative prior systems across six dimensions relevant to the failure modes discussed in Section~\ref{sec:related}. To the best of our knowledge, this is the first work to unify (1) deterministic symbolic verification of clinical intake completeness, (2) semantic entropy-based uncertainty quantification for multi-diagnosis outputs, and (3) recursive safety supervision within an open-source multi-agent architecture.

\section{Background and Task formulation}
\label{sec:methodology}

This section formalizes the clinical diagnosis task considered in this work. We first define the problem setting, and then describe the structured intake constraint, the uncertainty quantification mechanism, and the experimental design used for evaluation.

\subsection{Problem Formulation}
Let $\mathcal{C}$ denote the patient's chief complaint, and let $\mathcal{H} = \{h_1, h_2, \ldots, h_n\}$ denote the set of symptom details collected during the triage conversation. A clinical triage session is represented as a tuple $(\mathcal{C}, \mathcal{H}, \mathcal{D}, \mathcal{R})$, where $\mathcal{D}$ denotes the differential diagnosis generated by the diagnostic agent, and $\mathcal{R} \in \{0,1\}$ denotes the binary correctness label assigned by a ClinicalJudge agent.

The objective of the framework is to maximize the expected diagnostic correctness, $\mathbb{E}[\mathcal{R}]$, while ensuring that diagnosis is performed only after sufficiently complete symptom collection. To formalize this requirement, we define an OLDCARTS state vector
\[
V = \{V_{\text{Onset}}, V_{\text{Location}}, V_{\text{Duration}}, \dots, V_{\text{Severity}}\},
\]
where each $V_i \in \{0,1\}$ indicates whether the corresponding symptom dimension has been collected. The completeness condition is enforced through the constraint
\[
\sum_{i=1}^{8} V_i = 8,
\]
which requires that all eight OLDCARTS fields be observed before the system is allowed to transition from triage to diagnosis. In this setting, the problem is to design a triage-and-diagnosis workflow that jointly satisfies two requirements: (i) collect a complete and structured symptom history, and (ii) generate a diagnosis with high expected correctness.

\subsection{OLDCARTS Clinical Protocol}
The OLDCARTS mnemonic~\cite{bickley2012bates} specifies eight standard dimensions for structured symptom characterization in clinical history taking: Onset, Location, Duration, Character, Aggravating/Alleviating factors, Radiation, Timing, and Severity. To encode this protocol, we associate each dimension with a Boolean variable $V_i \in \{0,1\}$, initialized to $0$ and set to $1$ once the corresponding symptom attribute has been explicitly elicited and registered by the TriageNurse agent.

We define the intake completeness score as
\begin{equation}
    \sigma = \sum_{i=1}^{8} V_i, \quad \sigma \in \{0,1,\ldots,8\},
    \label{eq:sigma}
\end{equation}
where larger values of $\sigma$ indicate more complete symptom collection. A score of $\sigma=8$ corresponds to full coverage of all OLDCARTS dimensions. The M1 verification gate enforces the hard constraint $\sigma=8$ before the workflow is allowed to transition from triage to the diagnostic phase.
If a patient reports abdominal pain and the system collects only onset, location, and severity, then the corresponding state vector satisfies $\sigma=3$. In this case, the M1 gate blocks diagnostic handoff and requires the remaining OLDCARTS dimensions to be gathered before diagnosis can proceed.

\subsection{Semantic Entropy Uncertainty Quantification}
Following Kuhn et al.~\cite{kuhn2023semantic}, we adapt the Semantic Entropy (SE) estimator to the differential diagnosis setting. Let $\{s_1, s_2, \ldots, s_K\}$ denote $K$ independent diagnostic samples generated for the same clinical case. Each sample produces a set of candidate diagnoses, denoted by $L_i = \{d_1^{(i)}, d_2^{(i)}, \ldots\}$.

Because diagnostically equivalent outputs may appear in different lexical forms, we group semantically equivalent labels using bidirectional natural language inference (NLI). Specifically, two labels $a$ and $b$ are assigned to the same cluster $c_j$ if and only if
\begin{equation}
    P(\text{ENT} \mid a \to b) \geq \tau_{\text{NLI}} \;\wedge\; P(\text{ENT} \mid b \to a) \geq \tau_{\text{NLI}},
    \label{eq:nli}
\end{equation}
where $\tau_{\text{NLI}} = 0.7$. This clustering step reduces sensitivity to surface-level variation in diagnosis wording.

After clustering, we compute the normalized Shannon entropy over the cluster distribution:
\begin{equation}
    H = -\frac{1}{\log_2 |C|} \sum_{j=1}^{|C|} p_j \log_2 p_j, \quad H \in [0,1],
    \label{eq:entropy}
\end{equation}
where $p_j = |c_j|/n$, $|C|$ is the number of semantic clusters, and $n$ is the total number of diagnosis labels aggregated across all $K$ samples. A value of $H=0$ indicates complete agreement among the sampled outputs, whereas larger values of $H$ indicate greater disagreement and hence higher epistemic uncertainty.

We use a routing threshold $\tau = 0.5$. When $H > \tau$, the case is flagged to the MedicalSafetySupervisor together with an explicit uncertainty annotation, allowing uncertain diagnostic outputs to receive additional review before being presented to the user.

\noindent\textbf{Example.}  
Suppose the sampled outputs include labels such as \textit{myocardial infarction}, \textit{heart attack}, and \textit{gastric reflux}. The first two may be grouped into the same semantic cluster, while the third forms a different cluster. If the sampled diagnoses are distributed across multiple such clusters, the resulting entropy increases, indicating greater diagnostic disagreement.

\subsection{Ablation Study Design}
We conduct an ablation study to isolate the effects of the two proposed components: the OLDCARTS verification gate (M1) and the semantic entropy-based uncertainty quantification module (M2). The experimental conditions are defined by the Boolean setting of \texttt{USE\_OLDCARTS\_GATE} and the number of diagnostic samples $K$.

\begin{itemize}
    \item \textbf{Baseline (B):} Gate$=$F, $K=1$.  
    No symbolic intake verification is applied, and diagnosis is generated from a single deterministic sample.

    \item \textbf{Ablation A (A):} Gate$=$T, $K=1$.  
    The symbolic intake gate is enabled, but diagnosis is still based on a single deterministic sample; thus, semantic entropy-based uncertainty quantification is not used.

    \item \textbf{Full Architecture (FA):} Gate$=$T, $K=5$.  
    Both the symbolic intake gate and the uncertainty quantification module are enabled, allowing diagnosis to benefit from structured intake verification as well as multi-sample uncertainty estimation.
\end{itemize}

All three conditions are evaluated on $N \in \{50, 100, 150\}$ test cases, with $N=150$ used as the primary reporting setting. Diagnostic accuracy is determined by a ClinicalJudge agent, which assigns a binary score of $1$ when the ground-truth diagnosis matches any label in the pooled prediction set, and $0$ otherwise. For the single-sample settings ($K=1$), semantic entropy is undefined; in these cases, $H$ is set to $0.0$ as a placeholder, and no uncertainty-based routing is triggered.

\section{Design and Implementation}
\label{sec:design}

\subsection{Framework Overview}
Figure~\ref{fig:architecture} illustrates the proposed neuro-symbolic multi-agent framework for clinical triage and diagnosis. The framework is organized as a three-phase pipeline that combines structured symptom collection, uncertainty-aware diagnostic reasoning, and recursive safety supervision within a unified multi-agent workflow.

At the top of the pipeline, the \textit{User Interaction Layer} provides the entry point for patient input through a Gradio web interface. The patient supplies symptoms in natural language, and the interaction proceeds through a multi-turn conversation. These exchanges are forwarded to the \textit{Triage Nurse} agent, which is responsible for eliciting symptom details in a structured manner.

The first phase of the framework focuses on \textit{structured history taking}. During this phase, the Triage Nurse interacts with the patient and updates the \textit{shared conversation thread}, which serves as the central memory for downstream agents. A neuro-symbolic \textit{OLDCARTS gate} continuously verifies whether the required symptom dimensions:Onset, Location, Duration, Character, Aggravating/Alleviating factors, Radiation, Timing, and Severity; have been collected. If the completeness condition is not satisfied, the workflow is routed back to the Triage Nurse for additional questioning. Only when the OLDCARTS gate is satisfied is the case allowed to proceed to the diagnostic phase. In this way, the framework prevents premature transition to diagnosis based on incomplete clinical intake.

The second phase performs \textit{uncertainty-aware diagnostic reasoning}. Once the conversation history is deemed complete, the shared conversation thread is provided to a set of parallel \textit{Senior Diagnostician} agents. As shown in the figure, $K=5$ diagnostician instances are sampled at temperature $T=0.7$ in order to generate diverse candidate diagnoses for the same clinical case. Their outputs are then passed to an \textit{NLI clustering and entropy calculation} module, which groups semantically equivalent diagnosis labels and computes the semantic entropy score $H$. This score quantifies the level of disagreement across the sampled diagnostic outputs. Lower entropy indicates stronger agreement, whereas higher entropy reflects greater epistemic uncertainty.

The third phase performs \textit{deterministic diagnosis and safety review}. In parallel with the uncertainty estimation stage, a separate \textit{Senior Diagnostician} operating at temperature $T=0.0$ generates the final deterministic diagnostic report from the same shared conversation history. Before finalization, this diagnostician can invoke the \texttt{check\_drug\_interaction()} tool, which provides a static prototype mechanism for screening potentially unsafe medication combinations. The resulting diagnosis is then sent to the \textit{Medical Safety Supervisor}, which serves as the final review component in the framework.

The decision logic of the safety supervisor depends on both the generated diagnosis and the uncertainty signal. If the entropy score satisfies $H \leq \tau$, the diagnosis can proceed to final review under normal conditions. If $H > \tau$, the case is treated as uncertain and is subjected to closer supervision. The Medical Safety Supervisor either approves the diagnosis or rejects it and returns structured feedback. In the rejection case, the framework enters a \textit{recursive review loop}, shown in the figure by the feedback path from the supervisor back to the diagnostician, with a maximum of three revision attempts. This loop enables iterative refinement before an approved diagnosis is returned through the interface.

Overall, the framework combines symbolic verification and probabilistic reasoning in a coordinated multi-agent design. The OLDCARTS gate enforces completeness during intake, the semantic entropy module quantifies diagnostic uncertainty across multiple samples, and the safety supervisor provides an explicit review mechanism before patient-facing output is approved. This design allows the system to address both premature diagnostic handoff and silent hallucinations within a single end-to-end architecture.

\begin{figure*}[]
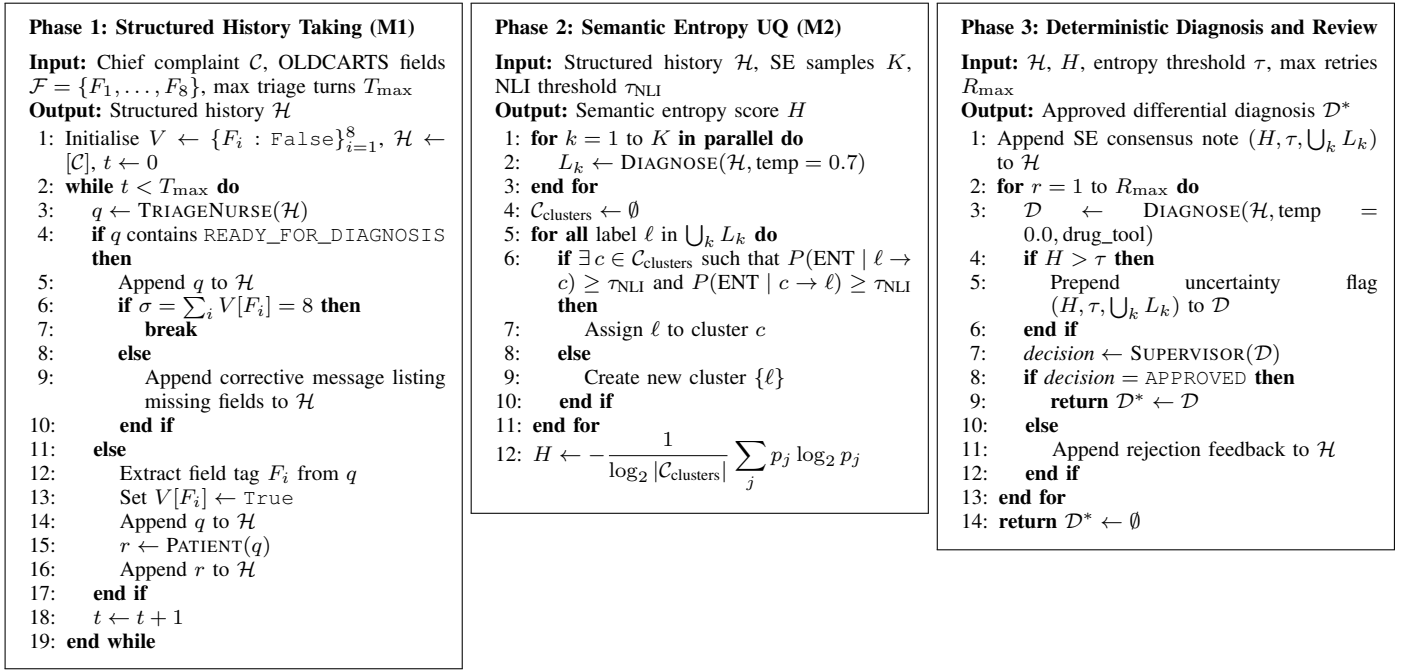

\centering
\setlength{\fboxsep}{6pt}

\begin{minipage}[t]{0.32\textwidth}
\footnotesize
\fbox{
\parbox[t]{0.95\linewidth}{
\textbf{Phase 1: Structured History Taking (M1)}\\[4pt]
\textbf{Input:} Chief complaint $\mathcal{C}$, OLDCARTS fields 
$\mathcal{F}=\{F_1,\ldots,F_8\}$, max triage turns $T_{\max}$\\
\textbf{Output:} Structured history $\mathcal{H}$

\begin{algorithmic}[1]
\STATE Initialise $V \leftarrow \{F_i:\texttt{False}\}_{i=1}^{8}$,
$\mathcal{H} \leftarrow [\mathcal{C}]$, $t \leftarrow 0$
\WHILE{$t < T_{\max}$}
    \STATE $q \leftarrow \textsc{TriageNurse}(\mathcal{H})$
    \IF{$q$ contains \texttt{READY\_FOR\_DIAGNOSIS}}
        \STATE Append $q$ to $\mathcal{H}$
        \IF{$\sigma = \sum_i V[F_i] = 8$}
            \STATE \textbf{break}
        \ELSE
            \STATE Append corrective message listing missing fields to $\mathcal{H}$
        \ENDIF
    \ELSE
        \STATE Extract field tag $F_i$ from $q$
        \STATE Set $V[F_i] \leftarrow \texttt{True}$
        \STATE Append $q$ to $\mathcal{H}$
        \STATE $r \leftarrow \textsc{Patient}(q)$
        \STATE Append $r$ to $\mathcal{H}$
    \ENDIF
    \STATE $t \leftarrow t + 1$
\ENDWHILE
\end{algorithmic}
}}
\end{minipage}
\hfill
\begin{minipage}[t]{0.32\textwidth}
\footnotesize
\fbox{
\parbox[t]{0.95\linewidth}{
\textbf{Phase 2: Semantic Entropy UQ (M2)}\\[4pt]
\textbf{Input:} Structured history $\mathcal{H}$, SE samples $K$, NLI threshold $\tau_{\text{NLI}}$\\
\textbf{Output:} Semantic entropy score $H$

\begin{algorithmic}[1]
\FOR{$k = 1$ to $K$ \textbf{in parallel}}
    \STATE $L_k \leftarrow \textsc{Diagnose}(\mathcal{H}, \text{temp}=0.7)$
\ENDFOR
\STATE $\mathcal{C}_{\text{clusters}} \leftarrow \emptyset$
\FORALL{label $\ell$ in $\bigcup_k L_k$}
    \IF{$\exists\, c \in \mathcal{C}_{\text{clusters}}$ such that
    $P(\text{ENT}\mid \ell \to c) \geq \tau_{\text{NLI}}$
    and
    $P(\text{ENT}\mid c \to \ell) \geq \tau_{\text{NLI}}$}
        \STATE Assign $\ell$ to cluster $c$
    \ELSE
        \STATE Create new cluster $\{\ell\}$
    \ENDIF
\ENDFOR
\STATE $H \leftarrow -\dfrac{1}{\log_2 |\mathcal{C}_{\text{clusters}}|}
\displaystyle\sum_j p_j \log_2 p_j$
\end{algorithmic}
}}
\end{minipage}
\hfill
\begin{minipage}[t]{0.32\textwidth}
\footnotesize
\fbox{
\parbox[t]{0.95\linewidth}{
\textbf{Phase 3: Deterministic Diagnosis and Review}\\[4pt]
\textbf{Input:} $\mathcal{H}$, $H$, entropy threshold $\tau$, max retries $R_{\max}$\\
\textbf{Output:} Approved differential diagnosis $\mathcal{D}^*$

\begin{algorithmic}[1]
\STATE Append SE consensus note $(H,\tau,\bigcup_k L_k)$ to $\mathcal{H}$
\FOR{$r = 1$ to $R_{\max}$}
    \STATE $\mathcal{D} \leftarrow \textsc{Diagnose}(\mathcal{H}, \text{temp}=0.0,\text{drug\_tool})$
    \IF{$H > \tau$}
        \STATE Prepend uncertainty flag $(H,\tau,\bigcup_k L_k)$ to $\mathcal{D}$
    \ENDIF
    \STATE $\textit{decision} \leftarrow \textsc{Supervisor}(\mathcal{D})$
    \IF{$\textit{decision} = \texttt{APPROVED}$}
        \STATE \textbf{return} $\mathcal{D}^* \leftarrow \mathcal{D}$
    \ELSE
        \STATE Append rejection feedback to $\mathcal{H}$
    \ENDIF
\ENDFOR
\STATE \textbf{return} $\mathcal{D}^* \leftarrow \emptyset$
\end{algorithmic}
}}
\end{minipage}

\caption{Three-phase algorithmic view of the proposed neuro-symbolic multi-agent clinical triage framework. The workflow consists of structured history taking with OLDCARTS verification, semantic entropy-based uncertainty quantification, and deterministic diagnosis with recursive safety review.}
\label{fig:framework_phases}
\end{figure*}

Figure~\ref{fig:framework_phases} formalises the complete three-phase execution of the proposed framework. The corresponding algorithmic workflow is organized into three phases. Phase~1 captures the structured triage procedure with neuro-symbolic OLDCARTS verification. Phase~2 computes semantic entropy over multiple independently generated diagnostic samples to estimate uncertainty. Phase~3 performs deterministic diagnosis generation followed by safety supervision and, if necessary, recursive revision before final approval.

\subsection{Agent Roles and Prompt Design}
The framework employs three role-specialized agents that correspond to the major stages of the clinical workflow: structured symptom elicitation, diagnostic reasoning, and safety supervision. Rather than relying on a single conversational model to perform all functions, these roles are separated to enable clearer control over information flow and to support phase-specific constraints at the orchestration layer.

\subsubsection{TriageNurse}
The \textit{TriageNurse} agent is responsible for structured symptom elicitation during the intake phase. Its prompt is designed to support focused, turn-by-turn questioning aligned with the OLDCARTS protocol. To make the elicited information machine-verifiable, each question is associated with a field tag indicating the symptom dimension being queried. These tags are consumed by the orchestration layer to update the OLDCARTS state vector and to determine whether the intake process is complete. The agent proceeds one question at a time and signals readiness for diagnostic handoff only after completing the symptom-gathering stage. This design ensures that the triage interaction remains both conversational and formally traceable.

\subsubsection{SeniorDiagnostician}
The \textit{SeniorDiagnostician} agent is responsible for generating the diagnostic assessment from the completed conversation history. Its output is structured to support both clinical interpretability and downstream uncertainty analysis. In particular, the agent produces a diagnosis-oriented report that includes the differential diagnosis and associated clinical recommendations, together with a machine-readable set of diagnosis labels used in semantic entropy computation. The same role is instantiated under two settings within the framework: multiple stochastic instances are used for uncertainty estimation, while a separate deterministic instance is used to generate the final patient-facing report. The diagnostician is also connected to a lightweight drug-interaction checking tool that serves as a prototype safeguard for medication-related recommendations.

\subsubsection{MedicalSafetySupervisor}
The \textit{MedicalSafetySupervisor} agent acts as the final review component before a diagnosis is approved. It evaluates the diagnostic report for clinical safety and either accepts the report or returns structured feedback for revision. When the uncertainty score exceeds the predefined threshold, the supervisor additionally receives an uncertainty summary containing the entropy value and the set of candidate diagnoses observed across the sampled outputs. Importantly, this uncertainty signal is treated as contextual information rather than as an automatic rejection criterion. The supervisor is intended to reject outputs only on the basis of clinical safety concerns, thereby separating uncertainty awareness from the final safety decision.

\subsection{M1: Neuro-Symbolic OLDCARTS State Tracker}
The M1 component implements a neuro-symbolic verification mechanism for monitoring intake completeness during triage. It maintains a Boolean state vector over the eight OLDCARTS dimensions, where each entry indicates whether the corresponding symptom attribute has been explicitly elicited during the conversation. From this state vector, the framework computes the completeness score $\sigma$ defined in Eq.~\ref{eq:sigma}, checks whether full coverage has been achieved, and identifies any symptom dimensions that remain uncollected. To make the symbolic interface explicit, field registration is implemented through pattern-based parsing of the TriageNurse output. Specifically, the orchestration layer detects tags of the form \texttt{[FIELD:X]}, where \texttt{X} denotes one of the OLDCARTS dimensions, using the following case-insensitive regular expression:

\begin{verbatim}
_FIELD_TAG_RE = re.compile(
  r"\[FIELD\s*:\s*(Onset|Location|Duration|
    Character|Aggravating|Radiation|
    Timing|Severity)\]\s*",
  re.IGNORECASE)
\end{verbatim}

This parser provides the symbolic grounding required for deterministic state updates: once a valid tag is detected, the corresponding OLDCARTS variable is marked as observed. An important design choice is the use of \texttt{re.search} rather than \texttt{re.match}. In practice, LLM-generated questions do not always begin directly with the field marker; they may first include a short natural-language preamble, e.g., ``The pain is sharp. \texttt{[FIELD:Character]} How would you describe it?'' An anchored match would fail in such cases, preventing the state tracker from registering the intended field. By instead searching for the tag anywhere in the generated response, the framework preserves deterministic symbolic registration while remaining robust to minor variations in model output formatting.

To support deterministic state updates, the TriageNurse associates each symptom-oriented question with an explicit field tag indicating the OLDCARTS dimension being queried. These tags are parsed by the orchestration layer and mapped to the corresponding entries in the state vector. Importantly, the tag extraction procedure is designed to be robust to variations in model output formatting. In practice, the field marker may appear either at the beginning of a question or after a short natural-language lead-in. The state tracker therefore searches for the tag anywhere in the generated response, rather than assuming a fixed positional format. This design choice makes the verification mechanism less sensitive to stylistic variation in LLM outputs while preserving deterministic field registration.

The gate is applied whenever the triage agent signals readiness to proceed to diagnosis. If the completeness condition $\sigma = 8$ is satisfied, the workflow is allowed to transition to the diagnostic stage. Otherwise, the orchestration layer blocks the handoff and returns corrective feedback identifying the missing OLDCARTS dimensions. This feedback is incorporated back into the conversation context so that the TriageNurse can continue collecting the required symptom details. In the evaluation setting, a more directive form of corrective feedback is used to explicitly reinforce the required field-tag format, which helps prevent repeated premature handoff attempts.

Overall, the M1 state tracker converts structured intake verification into an explicit system-level constraint rather than a prompt-level preference. As a result, the decision to proceed to diagnosis depends on the observed completeness of the conversation history, not on whether the language model chooses to follow the intended questioning protocol.

\subsection{M2: Semantic Entropy UQ Gate}
The M2 component implements uncertainty estimation through multi-sample diagnostic generation followed by semantic clustering. Given a completed conversation history, the framework generates $K=5$ independent diagnostic samples, each instantiated with a fresh agent and an independent model client at temperature $0.7$. The key requirement in this stage is sampling independence: no state or conversational context is shared across these diagnostician instances beyond the common input history. This design ensures that the sampled outputs can be treated as independent draws conditioned on the same clinical context. In the deployed system, these samples are generated concurrently to reduce end-to-end latency, whereas in the evaluation pipeline they are executed sequentially due to external rate constraints.

For each sample, the diagnosis labels used for uncertainty estimation are extracted separately from the free-form reasoning text. Concretely, the model output is structured such that the final line contains a machine-readable list of diagnosis names, which is parsed independently from the preceding explanation. This separation is important because entropy should reflect disagreement in the diagnosis space rather than superficial linguistic variation in the generated rationale.

To account for synonymous or lexically varied diagnoses, the extracted labels are grouped using bidirectional natural language inference. Specifically, the framework employs the \texttt{cross-encoder/nli-deberta-v3-small} model~\cite{hedeberta2021} to determine whether two diagnosis expressions are semantically equivalent. Since the NLI model is designed for sentence-pair inference, each diagnosis label is first converted into a simple sentential form (e.g., ``The diagnosis is \{x\}.'') before comparison. The resulting similarity module is implemented in a modular manner, allowing the semantic matching function to be replaced without changing the surrounding entropy computation pipeline. After clustering, the normalized Shannon entropy over the cluster distribution is computed as described in Eq.~\ref{eq:entropy}, and the resulting score is used as the uncertainty signal for downstream review.

\subsection{Concurrency and Session Management}
The framework is designed to support interactive clinical dialogue while preserving a clear separation between user interaction and backend orchestration. The user interface is handled through Gradio, while the multi-agent workflow executes asynchronously in a dedicated background thread with its own event loop. This separation allows the system to maintain responsive user interaction during long-running agent operations, including multi-turn triage, multi-sample uncertainty estimation, and recursive safety review.

Communication between the interface layer and the orchestration layer is managed through asynchronous queues and event-based synchronization primitives. This design enables controlled exchange of user inputs and agent outputs without blocking the interface thread. In addition, the framework maintains explicit bookkeeping over the shared conversation history to avoid repeated reprocessing of previously consumed messages. Rather than passing the full dialogue context at each step, each agent receives only the portion of the shared history that has not yet been consumed in its own stage of the workflow. This strategy reduces redundant context accumulation and improves efficiency during prolonged multi-agent interactions.

\section{Performance Evaluation}
\label{sec:evaluation}

\subsection{Evaluation Methodology}
We evaluate the proposed framework using a fully automated pipeline built on a clinical vignette dataset derived from the held-out test split of the MedQA-USMLE-4-options benchmark~\cite{jin2021medqa}. To align the evaluation with the diagnostic objective of our system, we retain only cases corresponding to diagnostic reasoning tasks, specifically those whose question text contains phrases such as ``most likely diagnosis'' or ``best next step.'' This filtering ensures that the ground-truth output for each case is a named diagnosis compatible with label-space evaluation.

For each selected vignette, the chief complaint is extracted from the original MedQA case using \texttt{Llama-3.1-8B-Instant} through a zero-shot JSON-formatted prompt. This preprocessing step produces structured \texttt{chief\_complaint} and \texttt{hidden\_vignette} fields that are stored in the evaluation set. Importantly, this auxiliary model is used only for data preparation and does not participate in the triage or diagnosis stages of the proposed framework.

During evaluation, each test case is instantiated as a simulated patient with access to the hidden vignette. The interaction begins from the extracted chief complaint, after which the \textit{TriageNurse} agent conducts the structured interview and the OLDCARTS state tracker records the completeness score $\sigma$ throughout the triage process. Once triage is completed, or once the maximum number of triage turns is reached (\texttt{MAX\_TRIAGE\_TURNS} $=15$), the diagnostic phase is initiated and $K$ diagnostic samples are collected for uncertainty estimation.

In the deployed system, these $K=5$ samples are generated concurrently. 
Diagnostic correctness is then determined by a \textit{ClinicalJudge} agent, which assigns a binary score $\mathcal{R} \in \{0,1\}$ depending on whether the ground-truth diagnosis appears in the pooled prediction set. To reduce ambiguity in automated scoring, the judge is instructed to return only the token \texttt{1} or \texttt{0}, although occasional verbose outputs containing a digit remain a minor limitation of this evaluation setup.

All experiments are conducted using llama-3.1-70b-instruct served via NVIDIA NIM, which replaces the Groq-based production endpoint to avoid token-per-minute constraints during large-scale evaluation. For each test case, we record case-level outputs including the completeness score, the number of premature handoff attempts, the semantic entropy value, the ground-truth diagnosis, the consensus diagnosis labels, and the final correctness indicator. Statistical analysis is performed post hoc using Pearson correlation, Fisher z-transformation, and 95\% bootstrap confidence intervals computed from $B=2000$ resamples with random seed $42$.

The recursive \textit{MedicalSafetySupervisor} loop is part of the deployed framework but is excluded from the automated evaluation reported here. This choice allows us to isolate the contribution of the M1 and M2 mechanisms without conflating their effects with the corrective behavior of the supervisor. Accordingly, the reported results should be interpreted as a conservative estimate of the framework’s diagnostic performance in the absence of the final review stage. Assessing the incremental contribution of the safety supervisor is left for future work involving clinician-centered evaluation.

\subsection{Ablation Study Results}

\begin{figure*}[t]
    \centering
    \includegraphics[width=\textwidth]{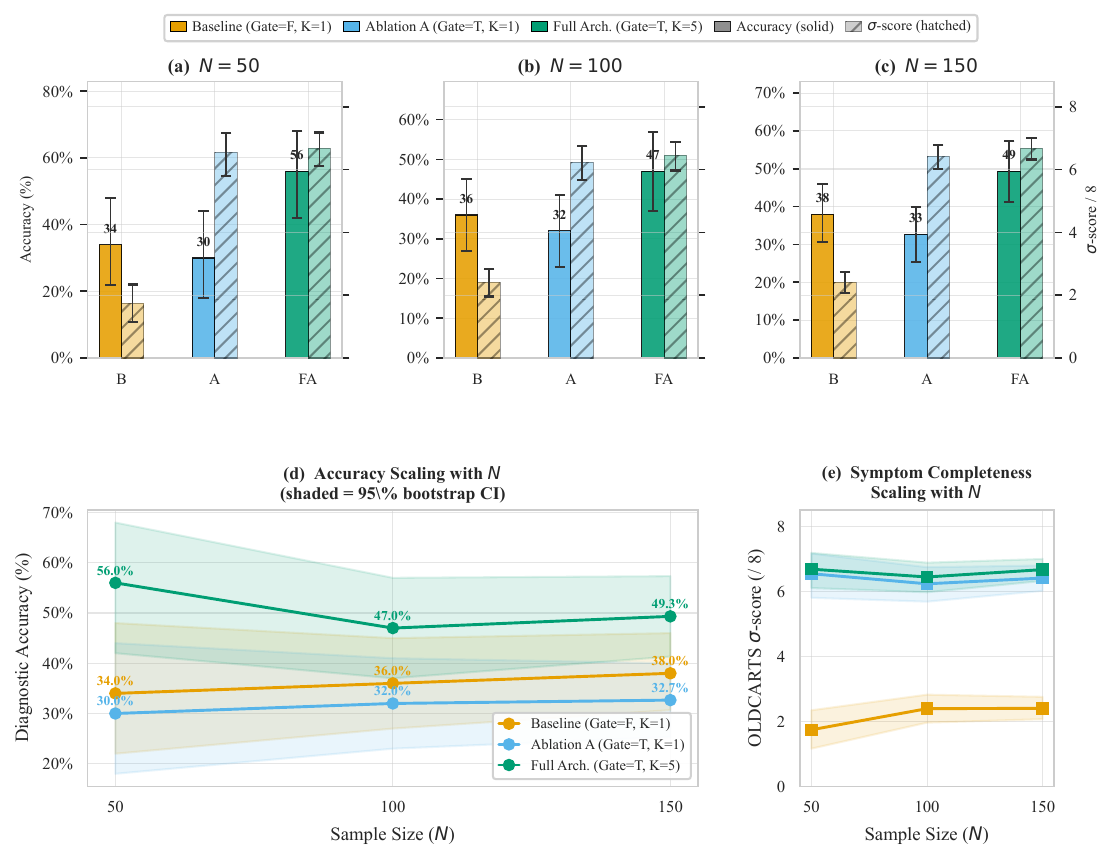}
    \caption{Ablation study across $N \in \{50, 100, 150\}$ test cases. Panels (a)--(c): diagnostic accuracy (solid bars) and OLDCARTS $\sigma$-score (hatched bars) with 95\% bootstrap CIs for Baseline (B), Ablation A (A), and Full Architecture (FA). Panels (d)--(e): accuracy and $\sigma$-score scaling with $N$ (shaded = 95\% bootstrap CI).}
    \label{fig:performance_scaling}
\end{figure*}

\begin{figure}[t]
    \centering
    \includegraphics[width=\columnwidth]{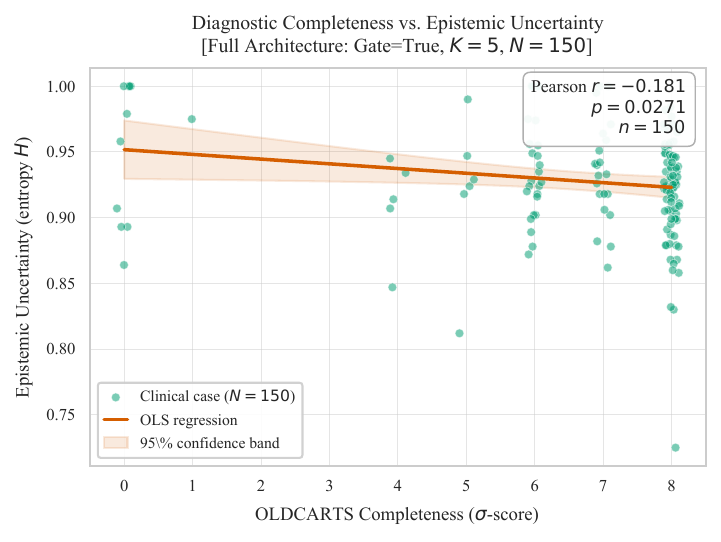}
    \caption{Linear regression analysis of the Full Architecture ($N=150$), demonstrating a statistically significant negative correlation ($r = -0.181$, $p = 0.0271$) between OLDCARTS completeness ($\sigma$-score) and downstream epistemic uncertainty ($H$). The OLS regression line with 95\% confidence band confirms the predicted direction: more complete symptom histories produce lower semantic entropy across $K=5$ independent diagnostic samples.}
    \label{fig:uncertainty_correlation}
\end{figure}

Table~\ref{tab:ablation} presents the mean $\sigma$-score and diagnostic accuracy for each ablation condition across $N \in \{50, 100, 150\}$ test cases. The corresponding scaling curves are shown in Fig.~\ref{fig:performance_scaling}.

\begin{table*}[t]
\centering
\caption{Ablation Study: Mean $\sigma$-score and Diagnostic Accuracy by Condition and Sample Size. Green rows = Full Architecture. $\Delta$ vs B = percentage-point difference vs Baseline. Mean $H$ reported for Full Architecture only.}
\label{tab:ablation}
\renewcommand{\arraystretch}{1.3}
\begin{tabular}{llccccc}
\toprule
\textbf{N} & \textbf{Condition} & \textbf{Gate} & \textbf{K} & \textbf{Mean $\sigma$-score} & \textbf{Accuracy (\%)} & \textbf{$\Delta$ vs B (pp)} \\
\midrule
\multirow{3}{*}{50}
  & Baseline (B)         & F & 1 & 1.740 & 34.0 & --- \\
  & Ablation A (A)       & T & 1 & 6.540 & 30.0 & $-4.0$ \\
  & \cellcolor{green!15}Full Arch. (FA) & \cellcolor{green!15}T & \cellcolor{green!15}5 & \cellcolor{green!15}6.680 & \cellcolor{green!15}\textbf{56.0} & \cellcolor{green!15}$+22.0$ \\
  & \multicolumn{5}{l}{\footnotesize Mean $H$ [FA]: 0.9342 \quad Pearson $r$ ($\sigma \Leftrightarrow H$) [FA]: $r = -0.0084$, $p = 0.954$ \quad 95\% CI: $[-0.286, 0.271]$} \\
\midrule
\multirow{3}{*}{100}
  & Baseline (B)         & F & 1 & 2.390 & 36.0 & --- \\
  & Ablation A (A)       & T & 1 & 6.230 & 32.0 & $-4.0$ \\
  & \cellcolor{green!15}Full Arch. (FA) & \cellcolor{green!15}T & \cellcolor{green!15}5 & \cellcolor{green!15}6.440 & \cellcolor{green!15}\textbf{47.0} & \cellcolor{green!15}$+11.0$ \\
  & \multicolumn{5}{l}{\footnotesize Mean $H$ [FA]: 0.9273 \quad Pearson $r$ ($\sigma \Leftrightarrow H$) [FA]: $r = -0.2703$, $p = 0.0065$ \quad 95\% CI: $[-0.443, -0.078]$} \\
\midrule
\multirow{3}{*}{150}
  & Baseline (B)         & F & 1 & 2.400 & 38.0 & --- \\
  & Ablation A (A)       & T & 1 & 6.407 & 32.7 & $-5.3$ \\
  & \cellcolor{green!15}Full Arch. (FA) & \cellcolor{green!15}T & \cellcolor{green!15}5 & \cellcolor{green!15}6.667 & \cellcolor{green!15}\textbf{49.3} & \cellcolor{green!15}$\mathbf{+11.3}$ \\
  & \multicolumn{5}{l}{\footnotesize Mean $H$ [FA]: 0.9278 \quad Pearson $r$ ($\sigma \Leftrightarrow H$) [FA]: $r = -0.1805$, $p = 0.0271$ \quad 95\% CI: $[-0.331, -0.021]$} \\
\bottomrule
\end{tabular}
\end{table*}

\subsection{OLDCARTS Completeness Analysis}


A direct consequence of disabling the M1 gate is a substantial reduction in the completeness of symptom collection. At $N=150$, the Baseline condition achieves a mean $\sigma$-score of only $2.400/8$, corresponding to 30.0\% completeness. This indicates that, without explicit intake verification, the TriageNurse collects on average fewer than three of the eight OLDCARTS dimensions before proceeding to diagnosis. These results provide quantitative evidence of the premature-handoff failure mode discussed in Section~\ref{sec:methodology}, where unconstrained conversational behavior does not reliably ensure structured clinical intake.

In contrast, the Full Architecture (Gate$=$T) attains a mean $\sigma$-score of $6.667/8$ at $N=150$, representing an absolute gain of $4.267$ points over the Baseline. This corresponds to an 83.3\% increase in average intake completeness. The remaining gap between $6.667$ and the maximum possible score of $8.0$ is primarily attributable to cases in which the triage process reaches the predefined limit of \texttt{MAX\_TRIAGE\_TURNS = 15} before all OLDCARTS dimensions are collected. This suggests that the residual incompleteness arises from turn-budget constraints rather than failure of the verification mechanism itself.

\subsection{Diagnostic Accuracy Analysis}

At $N=150$, the Baseline achieves a diagnostic accuracy of 38.0\%, whereas the Full Architecture reaches 49.3\%, yielding an absolute improvement of 11.3 percentage points. The largest gain is observed at $N=50$, where the Full Architecture achieves 56.0\% accuracy compared with 34.0\% for the Baseline, corresponding to a 22.0 percentage-point improvement. This pattern is consistent with the intuition that structured symptom collection is particularly beneficial when the available diagnostic context is otherwise limited.

Ablation A (Gate$=$T, $K=1$) underperforms the Baseline across all three evaluation settings: 30.0\% versus 34.0\% at $N=50$, 32.0\% versus 36.0\% at $N=100$, and 32.7\% versus 38.0\% at $N=150$. The reason is that activating the M1 gate increases the length of the triage history by preventing premature handoff, while the diagnosis stage in this setting still relies on a single deterministic sample without uncertainty-aware filtering. As a result, the additional context does not consistently translate into better predictions. In contrast, the Full Architecture compensates for this effect through multi-sample diagnostic generation and semantic clustering, which together recover and improve performance. These results suggest that the M1 and M2 components are most effective when deployed jointly, rather than in isolation.

\subsection{Statistical Analysis: $\sigma$-Score vs. Semantic Entropy $H$}

Fig.~\ref{fig:uncertainty_correlation} and the correlation statistics in Table~\ref{tab:ablation} address the central statistical hypothesis: \textbf{that higher OLDCARTS completeness ($\sigma$-score) is associated with lower epistemic uncertainty ($H$).} A complete symptom profile more strongly constrains the differential diagnosis space, producing greater consensus across $K$ independent samples and thus lower normalized Shannon entropy.

\begin{table}[t]
\centering
\caption{Pearson Correlation Between $\sigma$-Score and Semantic Entropy $H$ (Full Architecture Only). Bold row = primary reported result.}
\label{tab:correlation}
\renewcommand{\arraystretch}{1.3}
\begin{tabular}{ccccc}
\toprule
\textbf{N} & \textbf{$r$} & \textbf{$p$-value} & \textbf{95\% CI (Fisher $z$)} & \textbf{Sig.} \\
\midrule
50  & $-0.0084$ & $0.954$  & $[-0.286, 0.271]$   & No \\
100 & $-0.2703$ & $0.006$  & $[-0.443, -0.078]$  & Yes \\
\textbf{150} & $\mathbf{-0.1805}$ & $\mathbf{0.027}$ & $\mathbf{[-0.331, -0.021]}$ & \textbf{Yes} \\
\bottomrule
\end{tabular}
\end{table}

Table~\ref{tab:correlation} summarizes the correlation analysis. At $N=150$, the Pearson correlation between OLDCARTS completeness and semantic entropy is $r=-0.181$ ($p=0.027$, 95\% CI $[-0.331,-0.021]$), which is statistically significant at $\alpha=0.05$. The negative coefficient is consistent with the expected trend: cases with more complete symptom collection tend to exhibit lower semantic entropy across the $K=5$ diagnostic samples.

The observed effect size ($|r|=0.181$) is small in magnitude, which is reasonable in this setting. While the M1 gate directly affects the completeness of the collected clinical history, it does not directly constrain the semantic content of the downstream diagnosis. The relationship between $\sigma$ and $H$ is therefore indirect and mediated by the quality of the diagnostic reasoning process, which naturally introduces additional variability. At $N=50$, the correlation is not statistically significant ($r=-0.0084$, $p=0.954$), likely reflecting limited statistical power at the smaller sample size. A post-hoc power analysis based on the effect observed at $N=150$ suggests that approximately $N=120$ cases are required to achieve 80\% power at $\alpha=0.05$ (two-tailed), which is consistent with the emergence of statistical significance at $N=100$ and $N=150$.

\subsection{Scaling Analysis}

 Fig.~\ref{fig:performance_scaling}(e) plots diagnostic accuracy and OLDCARTS $\sigma$-score as a function of sample size. The accuracy of the Full Architecture is not monotonic with increasing $N$: it reaches 56.0\% at $N=50$ and then remains close to 49\% for $N \geq 100$. This pattern likely reflects variation in case difficulty at smaller sample sizes. Moreover, the overlapping 95\% bootstrap confidence intervals at $N=100$ and $N=150$ suggest that the difference between these two settings is not statistically significant. In contrast, the $\sigma$-scores for the gate-enabled conditions remain relatively stable across sample sizes (approximately $6.2$--$6.7$), indicating that the M1 mechanism enforces symptom completeness consistently. The Baseline $\sigma$-score is similarly stable, though at a much lower level (approximately $1.7$--$2.4$), showing that without explicit verification the model consistently fails to collect complete symptom histories regardless of sample size.

 Taken together, these results support three main observations:
\begin{enumerate}
    \item \textbf{M1 alone does not improve diagnostic accuracy:}  
    The results of Ablation A indicate that enforcing structured symptom completeness, by itself, does not lead to improved diagnostic performance. In fact, relative to the Baseline, accuracy decreases when the intake gate is enabled without accompanying uncertainty-aware filtering, suggesting that longer and more complete histories alone are not sufficient to improve prediction quality.

    \item \textbf{The benefits of M2 are realized most effectively in conjunction with M1:}  
    The accuracy gains of the Full Architecture arise from the combined use of structured intake verification and multi-sample uncertainty estimation. The M1 mechanism provides a more complete diagnostic context, while the M2 mechanism leverages this context to assess agreement across multiple samples and identify uncertain cases for additional scrutiny.

    \item \textbf{Evidence for the $\sigma$--$H$ relationship becomes clearer at larger sample sizes:}  
    The correlation between symptom completeness and semantic entropy does not reach statistical significance at $N=50$, but becomes significant for larger evaluation settings. This pattern is consistent with the sample sizes typically required to detect small correlation effects with adequate statistical power.
\end{enumerate}

\section{Conclusions and Future Work}
\label{sec:conclusions}

This paper proposed a neuro-symbolic multi-agent framework, based on Agentic AI paradigm, for clinical triage and diagnosis aimed at addressing two important failure modes in LLM-based healthcare agents: premature diagnostic handoff and silent hallucination. The framework combines a deterministic OLDCARTS-based intake verification mechanism with a semantic entropy-based uncertainty quantification module to enforce structured symptom collection and identify uncertain diagnostic outputs.

Experimental results show that the full architecture improves diagnostic accuracy by 11.3 percentage points over an unconstrained baseline at $N=150$. We also observe a statistically significant negative correlation between symptom completeness and diagnostic uncertainty, suggesting that more complete structured intake is associated with more consistent diagnostic predictions. Ablation results further indicate that the two components are most effective when used together.

While the current evaluation is based on automated clinical vignette testing, the proposed framework provides a foundation for more reliable multi-agent clinical decision support. Future work will focus on improving triage efficiency, tuning uncertainty thresholds more systematically, incorporating clinician-centered evaluation, and extending the framework to additional clinical modalities.

\bibliographystyle{IEEEtran}
\bibliography{references}

\end{document}